\pdfoutput=1

\documentclass[11pt]{article}

\usepackage{acl}

\usepackage{times}
\usepackage{latexsym}
\usepackage{pifont} %
\newcommand{\cmark}{\ding{51}}%
\newcommand{\xmark}{\ding{55}}%
\usepackage{comment}
\usepackage{graphicx}
\usepackage{booktabs}
\usepackage{hyperref}

\usepackage[T1]{fontenc}

\usepackage[utf8]{inputenc}

\usepackage{microtype}

\title{Few-Shot Document-Level Relation Extraction}

\author{Nicholas Popovic \and Michael Färber\\
  Karlsruhe Institute of Technology (KIT), Germany \\
  \texttt{\{popovic, michael.faerber\}@kit.edu} \\}

\begin{document}
\maketitle
\begin{abstract}
We present FREDo, a few-shot document-level relation extraction (FSDLRE) benchmark.
As opposed to existing benchmarks which are built on \textit{sentence-level} relation extraction corpora, we argue that \textit{document-level} corpora provide more realism, particularly regarding \textit{none-of-the-above} (NOTA) distributions.
Therefore, we propose a set of FSDLRE tasks and construct a benchmark based on two existing supervised learning data sets, DocRED and sciERC.
We adapt the state-of-the-art sentence-level method MNAV to the document-level and develop it further for improved domain adaptation.
We find FSDLRE to be a challenging setting with interesting new characteristics such as the ability to sample NOTA instances from the support set.
The data, code, and trained models are available online\footnote{\url{https://github.com/nicpopovic/FREDo}}.

\end{abstract}

\section{Introduction}

The goal of relation extraction is to detect and classify relations between entities in a text according to a predefined schema.
The schema, defining which relation types are relevant is highly dependent on the specific application and domain.
Supervised learning methods for relation extraction \cite{soares_matching_2019,zhou_document-level_2020,zhang_document-level_2021,xu_entity_2021,xiao_sais_2021}, which have advanced rapidly since the introduction of pretrained language models such as BERT \cite{devlin_bert_2019}, need large corpora of annotated relation instances to learn a schema.
Since annotating data sets for relation extraction manually is expensive and time consuming, few-shot learning for relation extraction represents a promising solution for relation extraction at scale. 

While the general $N$-way $K$-shot few-shot learning framework is relatively well defined and appears easy to apply to relation extraction, constructing realistic benchmark tasks has proven to be challenging. 
One of the core difficulties of establishing a realistic benchmark task for few-shot relation extraction is correctly modelling the most frequent situation a relation extraction system will encounter, \textit{none-of-the-above} (NOTA) detection.
NOTA refers to the case in which a candidate pair of entities does not hold any of the relations defined in the schema, a situation which is far more common than its reverse (for the document-level data set DocRED \cite{yao_docred_2019}, 96.84\% of candidate entity pairs are NOTA cases).
While initial benchmarks \cite{han_fewrel_2018} ignored this scenario altogether, researchers working on few-shot relation extraction have pushed for more realistic NOTA modeling in tasks and developed methods that can better detect NOTA instances \cite{gao_fewrel_2019,sabo_revisiting_2021}.

Parallel to the outlined efforts towards realistic few-shot relation extraction benchmarks, research into \textit{supervised} relation extraction has moved from \textit{sentence-level} tasks, relation extraction within single sentences, to \textit{document-level} relation extraction.
The push towards document-level relation extraction is motivated by (1) extracting more complex, cross-sentence relations and (2) information extraction at scale.
The latter is driven by an inherent challenge when increasing the scope from single sentences to multiple sentences: The number of entities involved increases and with that comes a quadratic increase in candidate entity pairs.
While sentence-level approaches typically evaluate each candidate entity pair individually, this strategy is infeasible at the document-level (DocRED contains an average of 393.5 candidate entity pairs per document, compared to only 2 for many sentence level-tasks).
In addition to the increased computation requirements, this results in a drastic increase in the amount of NOTA examples in a given query, demanding new methods of handling the imbalances that come with this change of distribution \cite{han_novel_2020,zhou_document-level_2020}.

All current few-shot relation extraction benchmarks are based on sentence-level tasks.
We argue that moving few-shot relation extraction from the sentence-level to the document-level: (1) brings with it as an inherent characteristic the more realistic NOTA distribution which prior work has looked to emulate and (2) will make the resulting methods more suitable for large scale information extraction.

In this work, we therefore define a new set of few-shot learning tasks for document-level relation extraction and design a strategy for creating realistic benchmarks from annotated document corpora.
Applying the above to the data sets DocRED \cite{yao_docred_2019} and sciERC \cite{luan_multi-task_2018}, we construct a few-shot document-level relation extraction (FSDLRE) benchmark, FREDo, consisting of two main tasks, an in-domain and a cross-domain task requiring domain adaptation. 
Finally, building on the state-of-the-art few-shot relation extraction approach MNAV \cite{sabo_revisiting_2021} and document-level relation extraction concepts \cite{zhou_document-level_2020}, we develop two approaches for tackling the above tasks.

We begin by outlining key related work in section \ref{sec:related_work}. %
In section \ref{sec:task} we give a description of the proposed tasks.
Next, in section \ref{sec:benchmark} we explain the construction of the benchmark, FREDo, followed by an overview of the proposed methods (section \ref{sec:experiments}), an analysis and discussion of the observed results (section \ref{sec:evaluation}), and the conclusion (section \ref{sec:conclusion}).

\section{Related Work}
\label{sec:related_work}

To the best of our knowledge, all current few-shot relation extraction benchmarks \cite{han_fewrel_2018, gao_fewrel_2019, sabo_revisiting_2021} focus on extracting relations from single sentences. 
FewRel \cite{han_fewrel_2018} introduces a relation extraction benchmark in the $N$-way $K$-shot setting \cite{vinyals_matching_2016,snell_prototypical_2017} in which a relation instance is to be assigned to one of $N$ classes given only $K$ examples for each of the classes.
In this setting human performance was quickly surpassed \cite{soares_matching_2019}, leading Gao et al. \cite{gao_fewrel_2019} to create FewRel 2.0 in an effort to increase the difficulty by adding a domain adaptation task, as well as a NOTA detection task. 
Sabo et al. \cite{sabo_revisiting_2021} argue that the way FewRel 2.0 models NOTA cases is not realistic due to the way NOTA instances are sampled, develop a framework for creating more realistic benchmarks and propose building such a benchmark using the sentence-level data set TACRED \cite{zhang_position-aware_2017}.
Tran et al. \cite{tran_one-shot_2021} forego labeled training data altogether and focus on the one-shot and weakly-supervised classification setting without NOTA cases.

While multiple relation extraction data sets based on annotated documents, rather than single sentences, are available in the form of CDR \cite{li_biocreative_2016}, sciERC \cite{luan_multi-task_2018}, SciREX \cite{jain-etal-2020-scirex}, DialogRE \cite{yu2020dialogue}, and GDA \cite{wu_renet_2019}, the introduction of the large scale data set DocRED \cite{yao_docred_2019} seems to have significantly increased research interest into supervised relation extraction at the document-level more recently \cite{zhou_document-level_2020,zhang_document-level_2021,xu_entity_2021,xiao_sais_2021}.

Since documents contain considerably more entities than individual sentences and the amount of candidate entity pairs increases quadratically with the amount of entities, applying sentence-level methods to document-level tasks is not feasible.
Document-level relation extraction approaches therefore use a different architecture \cite{wang_extracting_2019} than sentence-level approaches.
Another challenge is the large imbalance in the amount of positive and negative examples of relations encountered during training. 
Some researchers approach the problem by resampling training examples to counteract the imbalance \cite{han_novel_2020}, while others use more specialized solutions, such as modified loss functions \cite{zhou_document-level_2020}.

\section{Task Description}
\label{sec:task}

\begin{figure*}
\centering
    \includegraphics[width=0.9\textwidth]{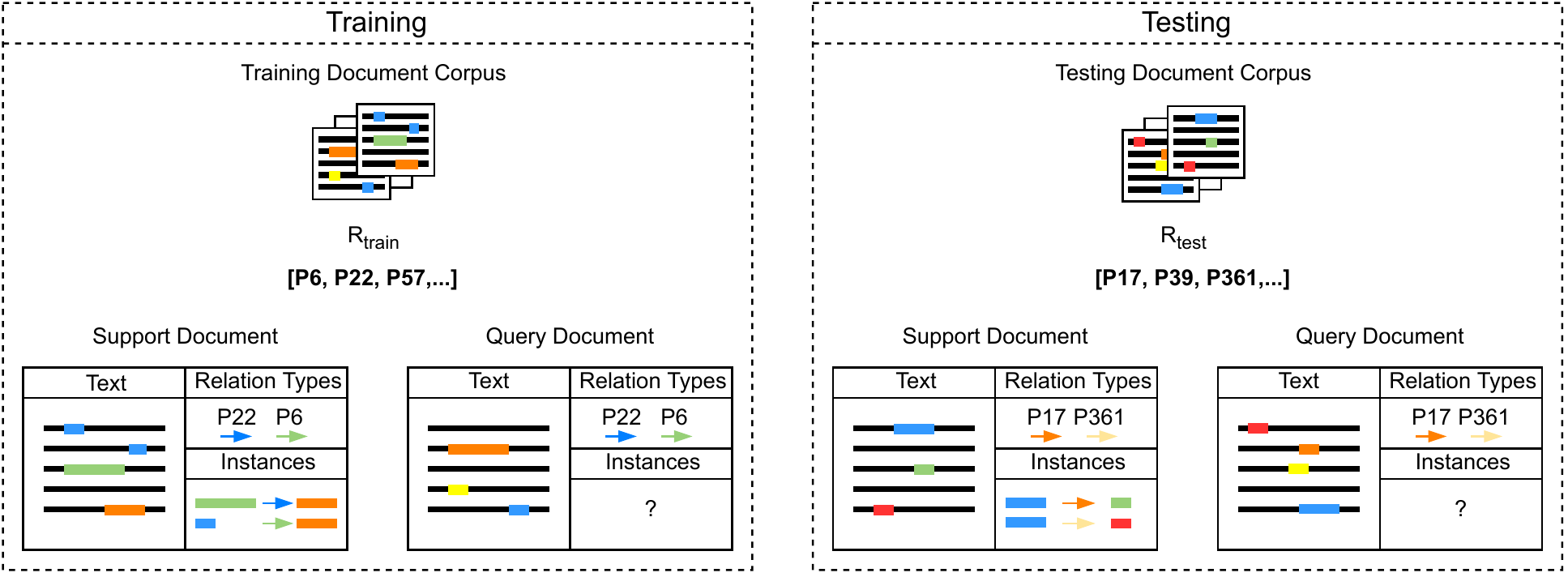}
    \caption{\label{fig:task} Illustration of an episode in the Few-Shot Document-Level Relation Extraction setting. Given a support document with annotated relation instances, the task is to return all instances of the same relation types for the query document. During testing a different corpus of documents, as well as a different set of relation types are used than during training.}
\end{figure*}

In document-level relation extraction the task is to return a set $S$ containing all valid triples of the format ($e_{h}$, $r_i$, $e_{t}$) for a document $D$.
Here, $e_{h}$ and $e_{t}$ are the head- and tail-entity of a relation instance, respectively, and
$r_{i} \in R$ is a relation type, with $R$ being the set of relation types for which instances are to be extracted.
The positions of any entity mentions, as well as any co-reference clusters are provided as part of the input\footnote{The setting in which no such annotations are given is typically referred to as \textit{joint entity and relation extraction} and is out of scope of this paper.}.
In both supervised learning and few-shot learning the documents used at test time are sampled from a different corpus than those used at training time.
The added complexity in few-shot learning is caused (1) by a change in the set of relation types $R$ between training and test time, and (2) by a much smaller amount of annotated examples given for each relation type.

\subsection{Document-Level Few-Shot Relation Extraction}
In figure \ref{fig:task} we give an illustration of the proposed task setting.
We define as \textit{few-shot} document-level relation extraction (FSDLRE) the following: 
Given a set of \textit{support documents} $\{D_{S,1}, ..., D_{S,k}\}$, the corresponding sets $\{T_{S,1}, ..., T_{S,k}\}$ containing \textit{all} valid triples for each support document, and a \textit{query document} $D_{Q}$, the task is to return the set $T_{Q}$, containing all valid triples in the query document.
The sets $\{T_{S,1}, ..., T_{S,k}\}$ and $T_{Q}$ consist of triples for the relation types $R_{episode}$. 
$R_{episode}$ is a subset of either $R_{train}$ or $R_{test}$, two disjoint sets of relation types used for training or testing, respectively.
The annotations of the support documents are complete, meaning that any candidate entity pair for which no relation type has been assigned can be considered NOTA.
\subsubsection{In-Domain vs. Cross-Domain}
For \textit{in-domain} FSDLRE training and test documents are taken from the same domain. 
For \textit{cross-domain} FSDLRE the test documents are taken from a different domain. 
Consequently, text style, text content, entity types, and relation types will all differ from those seen in the training documents. While this increases the difficulty of the challenge, this also resembles a more realistic application of few-shot relation extraction methods:
A key motivation for few-shot learning is to develop methods which can be applied to new data without the need for large-scale manual annotation. Restricting the applicability of a method to a specific domain and annotation procedure does not fit this idea.

\subsection{Differences to Existing Benchmarks}
The tasks described above differ from existing few-shot relation extracting benchmarks in multiple ways.
(1) Operating at document-level means the data now includes instances of relations expressed across multiple sentences and that models need to be able to evaluate candidate entity pairs more efficiently.
(2) Like for FS-TACRED \cite{sabo_revisiting_2021}, the amount of candidate entity pairs for which no relation is to be extracted is significantly larger than in other benchmarks (96.4\% compared to 15\%/50\% for FewRel 2.0 \cite{gao_fewrel_2019}) and the distribution from which NOTA instances are sampled, is more realistic than in FewRel 2.0, where NOTA instances are always instances of other, valid relation types.
(3) By requiring support annotations to be complete we have access to a support NOTA distribution, which is not the case for any existing benchmarks.
(4) Our tasks do not follow the fixed $N$-way $K$-shot format that related work has followed. Instead, $N$ and $K$ are variable between documents and therefore between individual episodes.

\section{FREDo: Few-Shot Document-Level Relation Extraction Benchmark}
\label{sec:benchmark}
\subsection{Selected Data Sets}

\begin{table*}
\centering
\renewcommand\thempfootnote{\arabic{mpfootnote}}
\begin{minipage}{\textwidth} %

\begin{tabular}{lcccccc}
\hline
\textbf{Data Set} & \textbf{\# Docs} & \textbf{\# RT\footnote{relation types}} & \textbf{\# CP/Doc\footnote{candidate pairs per document}} & \textbf{\# Words/Doc} & \textbf{\# Sents/Doc} & \textbf{Domain}\\
\hline
DocRED & 4051 & 96 & 394 & 172 & 8 & Non-specific\\
sciERC & 500 & 7 & 187 & 118 & 5.4 & Scientific Publications\\
\hline
\end{tabular}
\end{minipage}
\caption{\label{tab:data-statistics} A comparison of DocRED \cite{yao_docred_2019} and sciERC \cite{luan_multi-task_2018}, the datasets selected for the FREDo benchmark.}
\end{table*}

In order to construct a benchmark based on the tasks described in section \ref{sec:task} we require fully annotated data sets from two separate domains.
For the training set and the in-domain test set we use Doc\-RED \cite{yao_docred_2019} due to it being, to the best of our knowledge, the largest and most widely used document-level relation extraction data set. 
For the cross-domain test set we use sciERC \cite{luan_multi-task_2018} due to its domain (abstracts of scientific publications), which differs from DocRED (Wikipedia abstracts), and the fact that it contains annotations for 7 relation types. In table \ref{tab:data-statistics} we show a comparison of the selected datasets.
Additional document-level relation extraction data sets, SciREX \cite{jain-etal-2020-scirex}, DialogRE \cite{yu2020dialogue}, GDA \cite{wu_renet_2019}, CDR \cite{li_biocreative_2016}, were considered but ultimately not used for the cross-domain set, due to the amount of relation types annotated (too few), missing coreference links, or differing relation format (SciREX annotations are based on $N$-ary relations, while the other datasets annotate only binary relations).

\subsection{Training and Test Data}

\subsubsection{Document Corpora}
We begin by building 3 separate corpora of documents, 1 for training and development and 1 for testing each task (in-/cross-domain). Since the annotated test corpus for DocRED is not publicly available we use the documents in the development corpus as the test corpus for our in-domain task (meta-test). The DocRED training corpus is therefore used as the basis for both our training, and development set (meta-train). For the cross-domain task we require only a test set. This is because the training and development set for this task are identical to that of the in-domain task. We therefore use all documents in sciERC as our cross-domain test set (meta-test).
\subsubsection{Assigning Relation Types}
For preprocessing, we begin by comparing the relation types annotated in sciERC to those in the DocRED corpus\footnote{The mapping can be found in appendix \ref{sec:overlap}.}.
We find 2 relation types (P279, P361) which are annotated in both DocRED and sciERC. 
We remove these from the DocRED corpus in order to prevent data leakage between train and test sets. 

For DocRED, we split the remaining 94 relation types into 4 disjoint sets, a training set (62) $R_{train}$, development set (16) $R_{dev}$, and in-domain test set (16) $R_{test}$. 
For the cross-domain test set we use all 7 relation types in the sciERC corpus. 
An overview of the relation types assigned to each set can be found in appendix \ref{sec:overlap} and \ref{sec:types}.

\subsection{Test Episode Sampling}
In few-shot learning, each training/testing step consisting of support documents and query documents is called an episode.
Since evaluating every possible combination of support and query documents would result in too many episodes (approx. 1 million episodes for the in-domain and 250k episodes for the cross-domain test set) we need to sample a smaller amount of episodes from our corpora. 
We chose our sampling procedure with the goal of producing a representative measurement of the macro $F_1$ score.

For few-shot learning tasks the episode sampling process can be split into 2 steps, the first step being the sampling of the support examples and the second step being the sampling of the query examples. 
Unlike the sentence-level scenario where each example contains exactly one relation instance, each document we sample contains multiple instances of different relation types. 
In order to balance the amount of times each relation type is seen as a support example during testing we use the following procedures for the first sampling step: 
We begin by selecting from the set $R_{test}$ the relation type $r_s$ which is currently least represented in the testing corpus. 
If there are multiple such relation types we randomly choose one. 
For this relation type we sample support documents which contain at least one instance of $r_s$ each. 
Since the selected support documents may contain instances of other relation types in $R_{test}$ we add all of the relation types contained in the support document\footnote{The first, if there are multiple support documents.} to the episodes annotation schema. 
Following Sabo et al. \cite{sabo_revisiting_2021}, we randomly sample query documents from the test corpus\footnote{Note that we exclude the previously sampled support documents.} to realistically represent the NOTA distribution of the entire corpus.

\subsubsection{Choosing Test Set Sizes}

In order to choose a sufficiently large amount of test episodes for a representative $F_1$ score we evaluate a trained model for 50k episodes, logging the macro $f_1$ score at intervals of 100 episodes. 
We repeat this for 5 different random seeds. 
Using the variance between the 5 measurements as a guide, we choose a number of episodes which we deem to satisfy a good balance between low variance and manageable test set size. 
For robustness we sample episodes with 3 different random seeds for the final test sets.
The resulting test set sizes are: 15k episodes for the in-domain tasks and 3k episodes for the cross-domain tasks.

\subsection{Characteristics of Resulting Tasks}

\begin{table}
\centering
\begin{tabular}{lccc}
Task & $N$ & $K$ (micro) & $K$ (macro) \\
\hline
\multicolumn{4}{c}{in-domain}\\
\hline
1-Doc & 2.18 & 2.36 & 2.24 \\
3-Doc & 3.47 & 4.30 & 4.31 \\
\hline
\multicolumn{4}{c}{cross-domain}\\
\hline

1-Doc & 4.26 & 2.73 & 2.40 \\
3-Doc & 6.08 & 5.55 & 5.27 \\
\hline
\end{tabular}
\caption{\label{tab:n-k-dist-per-rt} Average values for $N$ and $K$ across test episodes. $K$ (micro) denotes the average across all episodes, $K$ (macro) denotes the weighted average of mean $K$ for each relation type.}
\end{table}

Existing few-shot benchmarks typically set 2 tasks, a single-shot and a $K$-shot (3/5/10-shot) challenge, in order to determine the way performance may scale when adding annotated training data.
Due to the nature of our tasks, $N$ and $K$ are variable from episode to episode, depending on the specific support documents and relation types.
We measure the scalability of approaches by defining a 1-Doc and 3-Doc challenge.

Therefore, the proposed benchmark, FREDo, consists of 2 main tasks with a 1-Doc and a 3-Doc subtask each:
\begin{itemize}
    \item The in-domain tasks for which an approach which has been trained on documents sampled from DocRED is evaluated on 15k episodes generated using documents from DocRED.
    \item The cross-domain tasks for which an approach which has been trained on documents sampled from DocRED is evaluated on 3k episodes generated using documents from sciERC.
\end{itemize}

In order to better characterize our tasks in relation to the common $N$-way $K$-shot format we measure the distribution of $N$ and $K$ across our test sets.
All the average values for $K$ and $N$ are shown in table \ref{tab:n-k-dist-per-rt}.
We find that the mean values of $N$ are (2.18/3.47) for the in-domain tasks (1-/3-doc) and (4.26/6.08) for the cross-domain task. For $K$ we calculate both the mean values across all episodes (micro), as well as the mean across the different relation types (macro).

\begin{figure*}
    \centering
    \scalebox{0.8}{\includegraphics{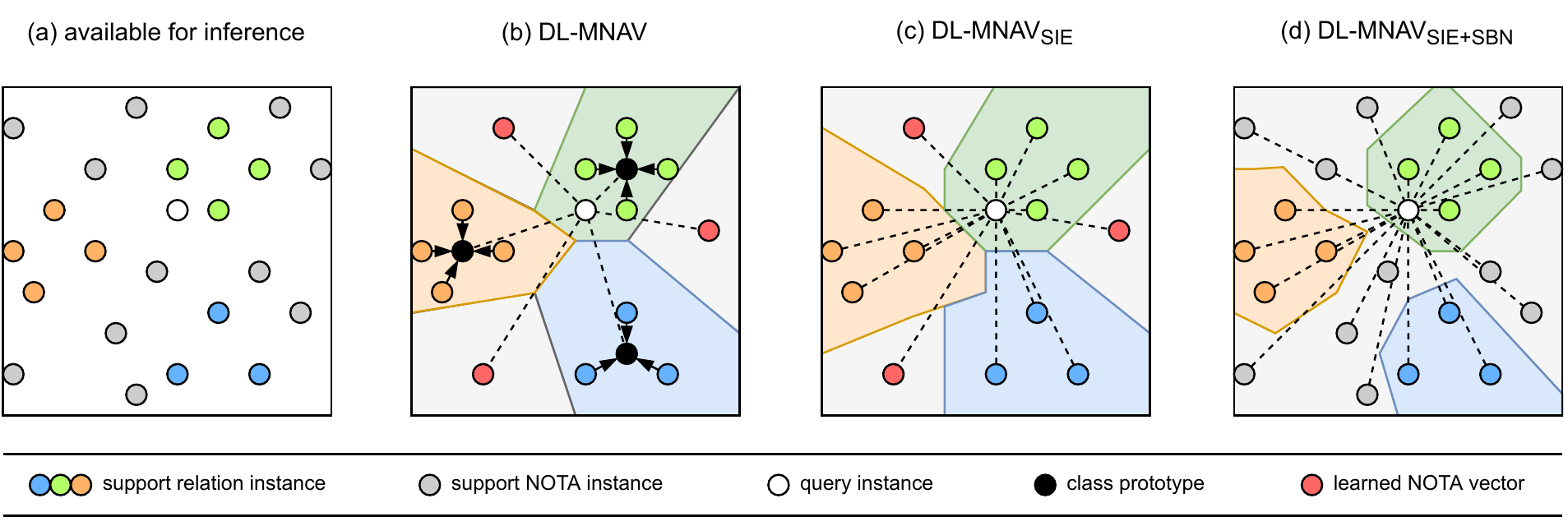}}
    \caption{Embedding space illustration of the different models used for evaluation \textit{at inference time}. Shown in (a) are all data points available from support documents, as well as the query instance. Class prototypes in (b) are computed as mean of support relation instances. Learned NOTA vectors, shown in red, are not based on support NOTA instances but learned during training and do not change based on the support document. Dotted lines indicate distances affecting the classification of the query instance, shown in white. Background colors illustrate approximate classification boundaries. For the baseline model (\ref{subsection:baseline}), decision boundaries are the same as in (d).}
    \label{fig:models}
\end{figure*}
\section{Experiments}
\label{sec:experiments}
\subsection{Models}

A common approach to relation extraction in general is to compute the similarity between embeddings produced by a fine-tuned language model such as BERT \cite{devlin_bert_2019}. 
In order to produce a relation embedding for a given pair of entities, most approaches concatenate embeddings corresponding to each entity.
One way to generate an entity embedding from the output of a language model is to average the embeddings of all tokens belonging to an entity.
Another way is the use of so called entity markers, introduced by Soares et al. \cite{soares_matching_2019}, which are tokens placed at the beginning and end of an entity mention within the input text.
The embeddings of the tokens placed at the start of each entity mention are then used as the entity embeddings.
In few-shot learning, a common way to use embedding similarity are prototypical networks \cite{snell_prototypical_2017}. 
Here, the embeddings of all $K$ support examples are averaged into a so-called prototype.
Given a query embedding, the similarity to the $N$ class-prototypes is then used for classification.

In order to assess the difficulty of our challenges we measure the performance of 3 approaches.
We set an initial baseline using the pretrained language model ${\textnormal{BERT}}_{\textnormal{BASE}}$ \cite{devlin_bert_2019} without fine-tuning.
Next, we adapt the state-of-the-art sentence-level few-shot relation extraction method MNAV \cite{sabo_revisiting_2021} to the document-level (DL-MNAV).
Finally, we make 2 modifications to DL-MNAV (SIE and SBN) to improve cross-domain performance.
In figure \ref{fig:models} we show a comparison of how the different models handle decision boundaries with respect to support and query instances.

\subsubsection{Baseline}
\label{subsection:baseline}
We set an initial baseline using the pretrained language model ${\textnormal{BERT}}_{\textnormal{BASE}}$ \cite{devlin_bert_2019} without fine-tuning in the following way: We encode each document using the language model and then average the output tokens of each entity mention. Following Han and Wang \cite{han_novel_2020}, we then average the mention representations for each entity. The resulting entity mentions are then concatenated for each candidate pair of entities to form relation embeddings. The similarity between a relation embedding in a query document to a relation embedding in a support document is calculated via their dot product. The relation type of the support embedding producing the highest dot product is output as the predicted relation type.

\subsubsection{Adapting MNAV to Document-Level (DL-MNAV)}
The current state-of-the-art few-shot sentence-level relation extraction method MNAV \cite{sabo_revisiting_2021} uses a combination of entity embeddings based on entity markers and prototypical networks.
Furthermore, it introduces the idea of learning $M$ prototypes to represent the NOTA class.
In order to use MNAV at the document-level one key architectural change is required: 
Instead of only marking two entities, a head- and a tail-entity, with two different tokens we mark all spans with the same entity marker tokens. 
Furthermore, following other document-level approaches \cite{han_novel_2020, zhou_document-level_2020} we apply a pooling step\footnote{For the pooling step we use mean pooling as, during preliminary experiments, it performed better than the logsumexp pooling used by Zhou et al..} to create representations for entities which are mentioned more than once in a document.

One of the challenges in realistic relation extraction is the large imbalance in the amount of positive and negative examples of relations encountered during training. 
In document-level relation extraction this challenge is even more central to the task than in sentence-level relation extraction. 
Preliminary experiments showed that simply using cross-entropy loss, as is done for MNAV, yields sub-par results.
To tackle this\footnote{We also examined the option of resampling the training samples such that the amount of negative examples in each training step roughly matches the amount of positive examples \cite{han_novel_2020}, but found this to be less effective.}, we adopt the adaptive thresholding loss function used by Zhou et al. \cite{zhou_document-level_2020} which is an adaptation of categorical cross entropy loss designed specifically for classifiers which treat NOTA as a relation type during classification, as is the case for MNAV.

Finally, we modify the initialization procedure of the NOTA vectors. While 
\citet{sabo_revisiting_2021} 
initialize the vectors using an averaged value of relation representations, we sample NOTA representations from the support documents during the first training step.

\subsubsection{Support Instance Evaluation (SIE)}
MNAV being based on prototypical networks \cite{snell_prototypical_2017} means that the embeddings of all support instances of a relation type are averaged into a single prototype.
While this has proven to be an effective strategy, we argue that it may not be ideal during inference in a cross-domain setting where the change in data distribution may break the assumption that the mean of support instances provides a good prototype.
In SIE we therefore use all individual support instances during inference, instead of their average.

\subsubsection{Support Based NOTA Vectors (SBN)}
In treating NOTA as a relation type and learning a persistent set of vectors during training, MNAV works on the assumption that the NOTA distribution during testing will match that seen during training. 
While this assumption is warranted and seems to work well for in-domain few-shot learning, we argue that this may not be the case for cross-domain settings. 
For this reason, we additionally add NOTA instances from the support documents to our set of NOTA vectors during training and inference.\footnote{The sampled NOTA representations do not persist across episodes.} 
Instead of randomly sampling NOTA vectors from the support documents we sample the most similar $k=5$ NOTA instances measured via their dot product for each relation prototype\footnote{Since, with SIE, we do not use prototypes at inference time, we then perform this sampling step for each relation instance rather than for prototypes and increase $k$ to 20.}.
During inference in a new domain, we use only the NOTA vectors sampled from the support document and ignore the learned vectors.

\subsection{Sampling Training \& Development Episodes}
\label{sec:train-sampling}
We compare 2 different ways of sampling episodes during training.
First we sample training and development episodes in the same way as for the test sets.
In order to get sufficient coverage to calculate representative macro $F_1$ scores on the development set, we sample 4k episodes.
As an alternative we modify the query sampling by ensuring that for each episode at least one of the query documents contains an instance of the relation type $r_s$.
This way we increase the amount of non-NOTA examples the model sees during training.
Another effect is that we need fewer development episodes (we use 500) to calculate macro $F_1$ scores.

\begin{table}
\centering
\begin{tabular}{lcc}
& \multicolumn{2}{c}{macro $F_1$ [\%]} \\
Model & 1-Doc & 3-Doc \\
\hline
Random Sampling & 5.77 & 5.29  \\
Ensure Positive & \textbf{7.26} & \textbf{9.37}  \\
\hline
\end{tabular}
\caption{\label{tab:results-sampling} Results for different query sampling strategies on the in-domain task}
\end{table}

\begin{table*}
\centering
\scalebox{0.88}{
\begin{tabular}{l@{\hspace{0.8\tabcolsep}}|ccc|ccc}
 & \multicolumn{3}{c|}{1-Doc} & \multicolumn{3}{c}{3-Doc} \\
Model & Precision [\%] & Recall [\%] & $F_1$ [\%] & Precision [\%] & Recall [\%] & $F_1$ [\%]\\
\hline
Baseline & 0.36 & 9.69 & 0.60 & 0.60 & 10.75 & 0.89  \\
DL-MNAV & $6.26 \pm 0.22$ & $21.08 \pm 2.71$ & $7.05 \pm 0.18$ & $7.71 \pm 0.69$ & $22.80 \pm 3.82$ & $\textbf{8.42} \pm \textbf{0.64}$  \\
$\textnormal{DL-MNAV}_{SIE}$ & $5.57 \pm 0.04$ & $23.12 \pm 1.69$ & $\textbf{7.06} \pm \textbf{0.15}$ & $5.16 \pm 0.19$ & $33.61 \pm 2.68$ & $6.77 \pm 0.21$ \\
$\textnormal{DL-MNAV}_{SIE+SBN}$ & $1.02 \pm 0.05$ & $22.94 \pm 1.87$ & $1.71 \pm 0.04$ & $1.75 \pm 0.16$ & $23.41 \pm 0.76$ & $2.79 \pm 0.24$ \\
\hline
\end{tabular}
}
\caption{\label{tab:results-in-domain} Results for FREDo in-domain task. Reported results are macro averages across relation types.}
\end{table*}

\begin{table*}
\centering
\scalebox{0.88}{
\begin{tabular}{l@{\hspace{0.8\tabcolsep}}|ccc|ccc}
 & \multicolumn{3}{c|}{1-Doc} & \multicolumn{3}{c}{3-Doc} \\
Model & Precision [\%] & Recall [\%] & $F_1$ [\%] & Precision [\%] & Recall [\%] & $F_1$ [\%]\\
\hline
Baseline & 1.34 & 3.04 & 1.76 & 1.84 & 2.47 & 1.98  \\
DL-MNAV & $2.30 \pm 0.45$ & $ 0.58 \pm 0.12$ & $0.84 \pm 0.16$ & $ 3.02 \pm 2.38$ & $ 0.29 \pm 0.13$ & $0.48 \pm 0.21$  \\
$\textnormal{DL-MNAV}_{SIE}$ & $ 1.77 \pm 0.60$ & $ 2.08 \pm 0.34$ & $1.77 \pm 0.60$ & $ 2.51 \pm 0.66$ & $ 2.52 \pm 0.31$ & $2.51 \pm 0.66$ \\
$\textnormal{DL-MNAV}_{SIE+SBN}$ & $ 2.26 \pm 0.11 $ & $ 4.37 \pm 0.13$ & $\textbf{2.85} \pm \textbf{0.12}$ & $ 3.47 \pm 0.14$ & $ 4.24 \pm 0.21$ & $\textbf{3.72} \pm \textbf{0.14}$ \\
\hline
\end{tabular}
}
\caption{\label{tab:results-cross-domain} Results for FREDo cross-domain task. Reported results are macro averages across relation types.}
\end{table*}

\section{Analysis of Results}
\label{sec:evaluation}
\subsection{Experimental Setup}
All our models are based on ${\textnormal{BERT}}_{\textnormal{BASE}}$ \cite{devlin_bert_2019} implemented using Huggingface's Transformers \cite{wolf_transformers_2020} and trained using mixed precision. We follow Zhou et al. \cite{zhou_document-level_2020} in using AdamW \cite{loshchilov_decoupled_2019} as optimizer (learning rates $\in [1\mathrm{e}{-5}, 3\mathrm{e}{-5}, 5\mathrm{e}{-5}, 1\mathrm{e}{-4}]$, of which $1\mathrm{e}{-5}$ generally performs best) and training using linear warmup (1k/2k steps) \cite{goyal_accurate_2017} followed by a linear learning rate decay. We use gradient clipping of $1.0$. We train each model for 50k episodes and perform early stopping based on the macro $F_1$ score on the development set which we measure every 1k/2k steps (when random sampling/ensuring positive examples). Each 1-doc training episode consists of 1 support document and 3 query documents, 3-doc training episodes contain 3 support documents and 1 query document. During training we limit the size of $R_{episode}$ to 1. We run each model 5 times using different random seeds and select the learning rate with the highest mean macro $F_1$ score on the development set for testing. For test scores we report the mean and standard deviation of macro $F_1$ scores for models trained using 5 different random seeds. For this model we report the macro $F_1$ score on the test set. Results are shown in tables \ref{tab:results-sampling}, \ref{tab:results-in-domain}, and \ref{tab:results-cross-domain}.
All models were trained on either NVIDIA V100 or NVIDIA 3090 GPUs.

\subsection{Comparing Sampling Strategies}

In table \ref{tab:results-sampling} we compare the test macro $F_1$ scores of the best models (chosen according to development set) trained using the 2 sampling strategies described in section \ref{sec:train-sampling} on the 1-Doc challenge using the model DL-MNAV. 
We find that ensuring positive query documents during training and development helps increase the $F_1$ scores.
Due to the better performance, as well as the reduced computation time needed for validation (thanks to the smaller development set), we use the latter strategy in all following experiments.

\subsection{Baseline Results}
As expected for a baseline which is not fine-tuned to the task at hand, the resulting macro $F_1$ scores are very low.
We argue, however, that the baseline is nevertheless relevant for 2 reasons. 
For the in-domain challenge, the baseline proves that the tasks are not trivially solvable by using a pre-trained language model out of the box.
For the cross-domain challenge, our baseline lets us see whether models overfit on the training domain.

\subsection{In-Domain Setting}

Test scores for the in-domain challenge are shown in table \ref{tab:results-in-domain}.
We observe large improvements in $F_1$ scores over the baseline, especially for DL-MNAV which reaches 7.05\% in the 1-Doc and 8.42\% in the 3-Doc task.
SIE does not seem to affect the accuracy of the model in the 1-Doc task; in the 3-Doc task, however, the $F_1$ score drops by 1.65 percentage points.
SBN, on the other hand causes the $F_1$ scores to drop by more than 5\%.
This performance drop following the removal of learned NOTA vectors clearly illustrates their effectiveness for in-domain tasks.

In table \ref{tab:benchmark_comparison} we compare the best $F_1$ scores of different few-shot relation extraction benchmarks.
Overall, compared to scores for benchmarks such as FewRel \cite{han_fewrel_2018} FewRel 2.0 \cite{gao_fewrel_2019}, the $F_1$ scores are considerably lower, illustrating the difficulty of such a realistic challenge.
When compared to the more realistic sentence-level benchmark FS-TACRED \cite{sabo_revisiting_2021} for which Sabo et al. report $F_1$\footnote{The reported results are micro $F_1$ scores} scores of 12.39\% (1-shot) and 30.04\% (5-shot) MNAV, these results are in-line with our expectations for an even more realistic (and thereby evidently more difficult) challenge.
Notably, the scaling behavior seen in FS-TACRED between the 1-shot and the 5-shot setting is not as pronounced for FREDo. We hypothesize that this is due the fact that the change in $K$ is not as large (see table \ref{tab:n-k-dist-per-rt}), meaning that (1) our 1-Doc setting does not correlate perfectly to the 1-shot setting, and (2) due to the way that additional support documents are sampled, the 3-Doc setting does not \textit{guarantee} additional support examples for infrequently occuring relation types.
\subsection{Cross-Domain Setting}

\begin{table}
\centering
\begin{tabular}{c@{\hspace{0.8\tabcolsep}}cccc}
Benchmark & input & realistic & best \\
 & length & NOTA & $F_1$ [\%] \\
\hline
FewRel & sentences & \xmark & 97.85\\
FewRel 2.0 & sentences & \xmark & 89.81\\
FS-TACRED & sentences & \cmark & 12.39\\
FREDo (ours) & documents & \cmark & 7.06\\
\hline
\end{tabular}
\caption{\label{tab:benchmark_comparison}A comparison highlighting the levels of difficulty of different few-shot relation extraction benchmarks. For all sentence-level benchmarks, we report the highest $F_1$ scores (at the time of writing) in the 5-way 1-shot setting. For FREDo we report the 1-Doc setting. For FewRel and FewRel 2.0, we report the highest scores found at the respective CodaLab competition websites.}
\end{table}

Test scores for the cross-domain challenge are shown in table \ref{tab:results-cross-domain}.
For DL-MNAV we see a significant drop in $F_1$ scores over the baseline, illustrating the problem with learned NOTA vectors in a cross-domain setting.
SIE brings the scores back to the baseline level, illustrating that the distribution of support examples is no longer well represented by their mean values.
Switching to SBN ($\textnormal{DL-MNAV}_{SIE+SBN}$), we find that our model exceeds the baseline scores, suggesting that the NOTA distribution on sciERC seems to be sufficiently different to cause an overfitting effect for learned NOTA vectors.
While SBN improves the results over the naive baseline, even the improved $F_1$ scores are extremely low. 
This is, however, unsurprising given the increase in difficulty over the previous setting.

\subsection{Scalability of DL-MNAV}
Although our methods show improvements over the proposed baseline in both tasks the results are currently severely lacking, especially compared to the state-of-the-art \textit{supervised} learning approaches on both data sets ($65.92\%$ $F_1$ for DocRED \cite{xu_entity_2021} and $52.0\%$ $F_1$ for sciERC \cite{ye_pack_2021}). This performance gap raises the question of whether our models will achieve similar performance if given enough support documents.
In order to assess the scalability of DL-MNAV when given amounts of annotated data resembling the supervised setting, we initialize a trained model with the full DocRED training corpus as support documents (96 classes, 3053 documents) and evaluate the performance on the full development set (96 classes, 998 documents). We measure an increase in recall to $45.75\%$ combined with a drop in precision to $5.75\%$, resulting in a $F_1$ score of $8.86\%$. 

While a direct comparison of this score with the few-shot settings assessed in FREDo is not appropriate, due to the nature of how the task is posed (different relation types are examined), the score can be compared to results obtained from supervised learning. 
Here we see clearly that, when trained in a few-shot setting, DL-MNAV does not scale well to the supervised setting.
We hypothesize that a key factor inhibiting the scalability is that the learned NOTA vectors are independent from the support documents in DL-MNAV.
As a result the NOTA representations of the model are unaffected by added support documents.
We do not experiment with SIE or SIE+SBN, as the amount of support instances would result in prohibitively large model sizes.\\

\subsection{Limitations}
Regarding the limitations of the proposed benchmark, FREDo, we believe that while it represents a good basis for model development, it will be beneficial to add other cross-domain data sets from a greater variety of domains in the future.
With the current, low $F_1$ scores seen in our tests, overestimating the performance of approaches does not seem to be too critical a danger. We are, however, hopeful that new methods might achieve significantly better results.
At that point we suggest a reassessment of how representative this benchmark is of cross-domain performance in general.
For the time being, however, we are confident that our tasks represent a valuable contribution to advancing the field.

\section{Conclusion}
\label{sec:conclusion}
In order to encourage the development of few-shot relation extraction approaches which are useful in real world scenarios, we propose FREDo, a few-shot document-level relation extraction benchmark.
By moving to the document-level, the settings become more realistic, a problem which existing benchmarks are struggling with.
For both in-domain and cross-domain tasks we present an approach which performs better than a simple baseline.
Our experiments confirm that, even though some existing benchmarks imply that impressive, even superhuman performance can already be achieved in few-shot relation extraction, realistic tasks are very difficult using current approaches and that significant advances are required for few-shot relation extraction approaches to be usable in a real word scenario.
In providing a benchmark that reveals this performance gap, we look to pave the way towards new methods with a potentially high impact on domain-specific and cross-domain relation extraction at scale.

\section*{Acknowledgements}

This work was partially supported by the German Federal Ministry of Education and Research (BMBF) as part of the project IIDI (01IS21026D) and the Smart Data Innovation Lab as part of the Smart Data Innovation Challenges (01IS19030A).
The authors acknowledge support by the state of Baden-Württemberg through bwHPC.

\bibliography{references}
\bibliographystyle{acl_natbib}

\appendix
\section{Overlap of DocRED and sciERC}
\label{sec:overlap}
\begin{table}
\centering
\begin{tabular}{lcc}
\hline
\textbf{sciERC ID} & \textbf{Wikidata ID} & \textbf{DocRED}\\
\hline
hyponym-of & P279 & \cmark\\
part-of & P361 & \cmark\\
used-for & P366 & \xmark\\
compare & P2210 & \xmark\\
evaluate-for & P5133 & \xmark\\
feature-of & - & -\\
conjunction & - & -\\
\hline
\end{tabular}
\caption{\label{tab:matched-types} Overlap of relation types in sciERC and Doc\-RED}
\end{table}

In table \ref{tab:matched-types} we show the mapping of sciERC relation types onto Wikidata \cite{vrandecic_wikidata_2014} properties and whether these relation types are contained in DocRED.

\section{Relation Types in in-domain dataset}
\label{sec:types}

In tables \ref{tab:train-types1}-\ref{tab:test-types}, we list the relation types in the different datasets based on DocRED.

\begin{table*}
\centering
\begin{tabular}{llr}
\hline
Wikidata ID & Description & Number of instances\\
\hline
P131 & located in the administrative territorial entity & 4193\\
P577 & publication date & 1142\\
P175 & performer & 1052\\
P569 & date of birth & 1044\\
P570 & date of death & 805\\
P161 & cast member & 621\\
P264 & record label & 583\\
P527 & has part & 632\\
P19 & place of birth & 511\\
P54 & member of sports team & 379\\
P40 & child & 360\\
P30 & continent & 356\\
P69 & educated at & 316\\
P26 & spouse & 303\\
P607 & conflict & 275\\
P159 & headquarters location & 264\\
P22 & father & 273\\
P400 & platform & 304\\
P1344 & participant of & 223\\
P206 & located in or next to body of water & 194\\
P127 & owned by & 208\\
P170 & creator & 231\\
P178 & developer & 238\\
P20 & place of death & 203\\
P1412 & languages spoken, written or signed & 155\\
P155 & follows & 188\\
P710 & participant & 191\\
P6 & head of government & 210\\
P108 & employer & 196\\
P276 & location & 172\\
P156 & followed by & 192\\
P166 & award received & 173\\
P123 & publisher & 172\\
P800 & notable work & 150\\
P449 & original network & 152\\
P58 & screenwriter & 156\\
P706 & located on terrain feature & 137\\
P162 & producer & 119\\
P37 & official language & 119\\
P241 & military branch & 108\\
P31 & instance of & 103\\
P403 & mouth of the watercourse & 95\\
P580 & start time & 110\\
P585 & point in time & 96\\
P749 & parent organization & 92\\
P937 & work location & 104\\
P36 & capital & 85\\
P576 & dissolved, abolished or demolished & 79\\
P172 & ethnic group & 79\\
P205 & basin country & 85\\
P1376 & capital of & 76\\

\hline
\end{tabular}
\caption{\label{tab:train-types1} Relation types present in training data (continued on next page).}
\end{table*}

\begin{table*}
\centering
\begin{tabular}{llr}
\hline
Wikidata ID & Description & Number of instances\\
\hline
P171 & parent taxon & 75\\
P740 & location of formation & 62\\
P840 & narrative location & 48\\
P676 & lyrics by & 36\\
P1336 & territory claimed by & 33\\
P551 & residence & 35\\
P1365 & replaces & 18\\
P737 & influenced by & 9\\
P190 & sister city & 4\\
P807 & separated from & 2\\
P1198 & unemployment rate & 2\\

\hline
\end{tabular}
\caption{\label{tab:train-types2} Relation types present in training data (continued).}
\end{table*}

\begin{table*}
\centering
\begin{tabular}{llr}
\hline
Wikidata ID & Description & Number of instances\\
\hline
P27 & country of citizenship & 2689\\
P150 & contains administrative territorial entity & 2004\\
P571 & inception & 475\\
P50 & author & 320\\
P1441 & present in work & 299\\
P57 & director & 246\\
P179 & series & 144\\
P137 & operator & 95\\
P112 & founded by & 100\\
P86 & composer & 79\\
P176 & manufacturer & 83\\
P355 & subsidiary & 92\\
P136 & genre & 111\\
P488 & chairperson & 63\\
P1366 & replaced by & 36\\
P1056 & product or material produced & 36\\
\hline
\end{tabular}
\caption{\label{tab:dev-types} Relation types present in development/validation data.}
\end{table*}

\begin{table*}
\centering
\begin{tabular}{llr}
\hline
Wikidata ID & Description & Number of instances\\
\hline
P17 & country & 2831\\
P361 & part of & 194\\
P495 & country of origin & 212\\
P102 & member of political party & 98\\
P463 & member of & 113\\
P3373 & sibling & 134\\
P1001 & applies to jurisdiction & 83\\
P118 & league & 56\\
P674 & characters & 74\\
P194 & legislative body & 56\\
P140 & religion & 82\\
P35 & head of state & 51\\
P364 & original language of work & 30\\
P272 & production company & 36\\
P279 & subclass of & 36\\
P25 & mother & 15\\
P582 & end time & 23\\
P39 & position held & 8\\
\hline
\end{tabular}
\caption{\label{tab:test-types} Relation types present in test data.}
\end{table*}

\end{document}